\documentclass[10pt,twocolumn,letterpaper]{article}

\usepackage{iccv}
\usepackage{times}
\usepackage{epsfig}
\usepackage{graphicx}
\usepackage{amsmath}
\usepackage{amssymb}
\usepackage[dvipsnames, svgnames]{xcolor}

\usepackage{placeins}
\usepackage{array}
\usepackage{enumerate}
\usepackage{algpseudocode}
\usepackage{algorithm}

\usepackage[pagebackref=true,breaklinks=true,letterpaper=true,colorlinks,bookmarks=false]{hyperref}

\newcolumntype{P}[1]{>{\centering\arraybackslash}p{#1}}

\newcommand{\vct}[1]{\mathbf{#1}}

\iccvfinalcopy 

\def\iccvPaperID{0000} 

\begin{document}

\title{Learnable Online Graph Representations for 3D Multi-Object Tracking}

\author{Jan-Nico~Zaech$^{1}$\quad Dengxin~Dai$^{1}$\quad Alexander Liniger$^{1}$\quad Martin Danelljan$^{1}$\quad Luc~Van~Gool$^{1,2}$\\
$^{1}$Computer Vision Laboratory, ETH Zurich, Switzerland, $^{2}$KU Leuven, Belgium\\
{\tt\small\{zaechj,dai,alex.liniger,martin.danelljan,vangool\}@vision.ee.ethz.ch}
}

\maketitle

\begin{abstract}
Tracking of objects in 3D is a fundamental task in computer vision that finds use in a wide range of applications such as autonomous driving, robotics or augmented reality. Most recent approaches for 3D multi object tracking (MOT) from LIDAR use object dynamics together with a set of handcrafted features to match detections of objects.
However, manually designing such features and heuristics is cumbersome and often leads to suboptimal performance. 
In this work, we instead strive towards a unified and learning based approach to the 3D MOT problem.
We design a graph structure to jointly process detection and track states in an online manner. To this end, we employ a Neural Message Passing network for data association that is fully trainable. Our approach provides a natural way for track initialization and handling of false positive detections, while significantly improving track stability.
We show the merit of the proposed approach on the publicly available nuScenes dataset by achieving state-of-the-art performance of 65.6\% AMOTA and 58\% fewer ID-switches.
\end{abstract}

\FloatBarrier
\section{Introduction}
\label{sec:intro}

\begin{figure}[t]
    \centering
    \includegraphics[width=\linewidth, trim=110 00 120 0, clip]{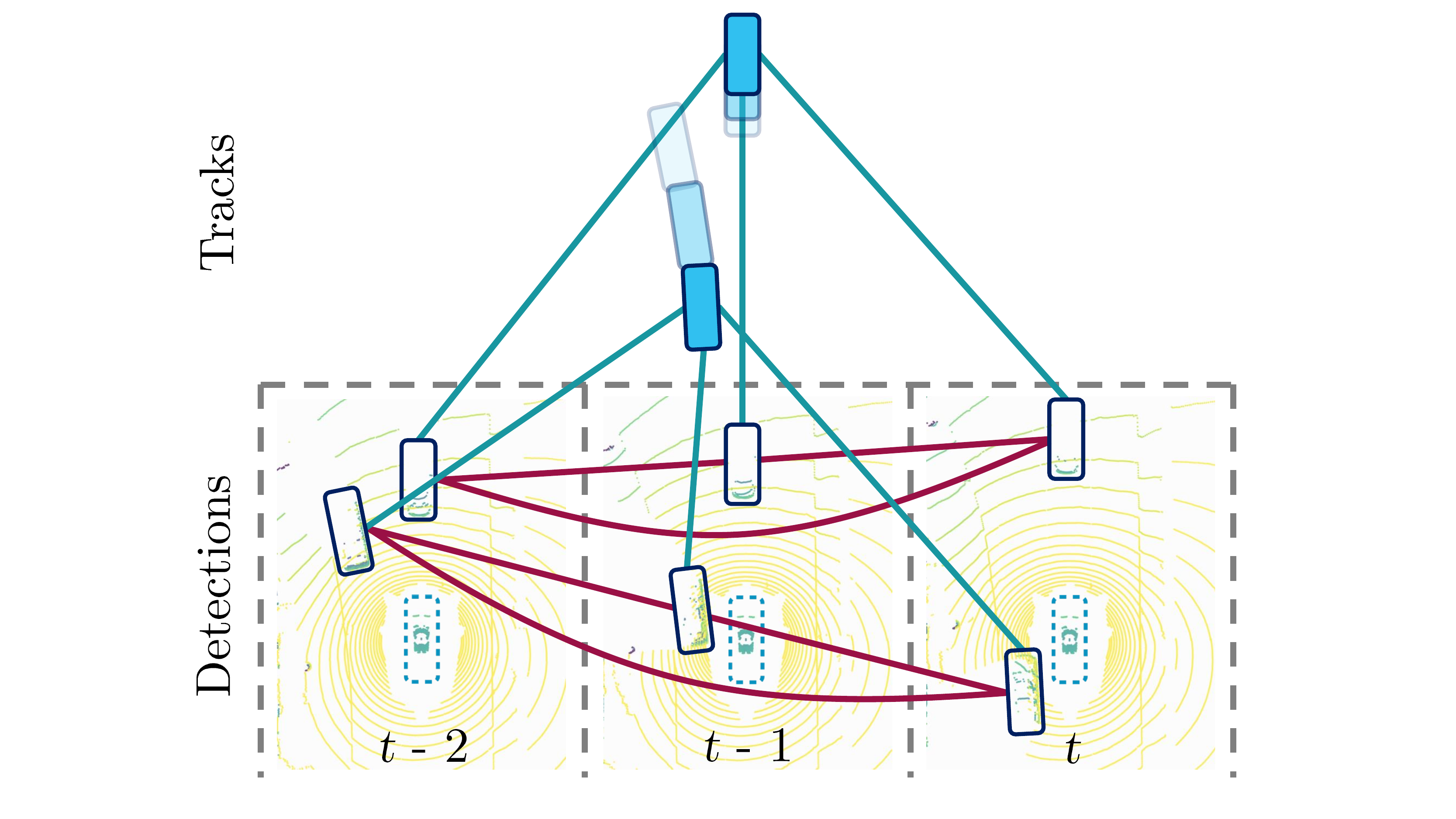}
    \caption{The proposed method uses a graph representation for detections and tracks. A neural message passing based architecture performs matching of detections and tracks and provides a learning based framework for track initialization, effectively replacing heuristics that are required in current approaches.}
    \label{fig:method_overview}
\end{figure}

Autonomous systems require a comprehensive understanding of their environment for a safe and efficient operation. A task at the core of this problem is the capability to robustly track objects in 3D in an online-setting, which enables further downstream tasks like path-planning and trajectory prediction~\cite{alahi_social_2016, Hong_2019_CVPR, zaech_action_2020}.
Nevertheless, tracking multiple objects in 3D in order to operate an autonomous system, poses major challenges.
First, in the online setting, data association, track initialization, and termination need to be solved under additional uncertainty, as only past and current observations can be utilized. Furthermore, covering occlusions requires extrapolation with a predictive model rather than interpolation as in the offline case. Finally, when using LIDAR for data acquisition, no comprehensive appearance data is available and data association needs to primarily rely on object dynamics. This is further complicated by the presence of fast moving objects such as cars.

With the release of large scale datasets for 3D tracking~\cite{nuscenes2019,Argoverse,kesten_lyft_2019,sun_scalability_2020}, a considerable amount of work on 3D MOT has been initiated~\cite{chiu_probabilistic_2020,kim_eagermot_2020,weng_gnn3dmot_2020,Weng-2020-123397,yin_centerbased_2020}. Most of these works address the aforementioned challenges by either linking detections directly in a learning based manner or use comprehensive motion models together with handcrafted matching metrics. All of these methods require a large set of heuristics and, to the best of our knowledge, none of the methods approaches the aforementioned challenges jointly.
In contrast to this, recent work in 2D MOT \cite{braso_learning_2020, bergmann_tracking_2019} aims at reducing the amount of heuristics by modeling all tasks in a single learnable pipeline using graph neural networks. However, most of these approaches are limited to the offline setting and driven by appearance-based association that cannot be readily employed in the 3D counterpart. 

To establish the missing link between learning based methods and powerful predictive models in 3D MOT, we propose a unified graph representation that merges tracks and their predictive models with object detections into a single graph. This learnable formulation effectively replaces heuristics that are required in current methods. A visualization of the graph is depicted in Figure \ref{fig:method_overview}.

Contrary to previous works, our learnable matching between tracks and detections is integrated into a closed-loop tracking pipeline, alleviating the need for handcrafted features. However, this raises the question of how to effectively train such a learnable system, as the generated tracks influence the data distribution seen during subsequent iterations. In this work, we propose a two-stage training procedure for semi-online training of the algorithm, where the data seen during training is generated by the model itself.
In summary, the contributions of our work are threefold:
\begin{itemize}
    \setlength{\itemsep}{0pt}
    \setlength{\parskip}{0pt}
    \item A unified graph representation for learnable online 3D MOT that jointly utilizes predictive models and object detection features.
    \item A track-detection association method that explicitly utilizes relational information between detections to further improve track stability.
    \item A training strategy that allows us to faithfully model online inference during learning itself.
\end{itemize}
We perform extensive experiments on the challenging nuScenes dataset. Our approach sets a new state-of-the-art, achieving an AMOTA score of 0.656 while reducing the number of ID-switches by 58\%.

\section{Related Work}
\label{sec:rel_work}

\paragraph*{2D MOT} is a well investigated task, with the MOT challenge~\cite{dendorfer_mot20_2020,leal-taixe_motchallenge_2015,milan_mot16_2016} and its corresponding dataset as the current performance reference. The general goal in 2D MOT is to detect and track known objects of a single type or multiple types.
A widely adopted approach to MOT is tracking by detection, where detections are available from an independently trained detection module and data association is performed by the tracker. Due to the nature of the task, a wide range of approaches cast tracking as a graph problem~\cite{braso_learning_2020,li_global_2008,roshanzamir_gmcptracker_2012,tang_multiple_2017}.

Following the paradigm of combining detection and tracking into a single module, Tracktor~\cite{bergmann_tracking_2019} uses the box regression module of faster RCNN~\cite{ren_faster_2017} to propagate and refine object bounding boxes between frames.
A range of tracker extensions are commonly used in all approaches, including modules such as camera motion compensation~\cite{bergmann_tracking_2019} or object re-identification (ReID)~\cite{karthik_simple_2020,ma_customized_2019}.
In general, most of the 2D MOT methods profit from the high framerate available in videos~\cite{bergmann_tracking_2019}. Furthermore, state-of-the-art 2D object detectors achieve a high accuracy~\cite{he_mask_2017,ren_faster_2017,tan_efficientdet_2020}, such that the focus of tracking has shifted from the rejection of false positives towards a pure data assignment task~\cite{braso_learning_2020}.

Closest to ours approach, NMPtrack~\cite{braso_learning_2020} introduces Neural Message Passing (NMP)
as a graph neural solver for offline 2D pedestrian tracking. Starting from a network flow formulation, the problem is transformed into a classification problem and data assignment is solved with NMP. Also using NMP as the network solver, we propose a graph representation that is capable of \textbf{online 3D tracking} and integrate a state filter for track representation. In contrast to~\cite{braso_learning_2020}, we do not require the complete sequence of frames to be available and do not assume that false positive detections are absent. Therefore, we are able to perform online tracking, while considering predictions in frame-gaps and taking imperfect object detectors into account.

\paragraph*{3D MOT} extends the challenge of MOT to tracking multiple objects in 3D~\cite{nuscenes2019, geiger_vision_2013}. With 3D MOT as a problem at the core of autonomous driving, a wide range of datasets that focus on tracking of objects in driving scenes is available~\cite{Argoverse,kesten_lyft_2019,nuscenes2019,sun_scalability_2020}. Due to the nature of the task, 3D MOT is usually performed online, which adds additional challenges and requires additional heuristics. For detecting objects, any 3D modality would be suitable, nevertheless, most datasets provide LIDAR scans which are used in most methods, including ours. As 3D object detection from LIDAR is still an open research question and less robust than 2D detection, 3DMOT mostly follows the tracking by detection framework~\cite{chiu_probabilistic_2020,kim_eagermot_2020,weng_gnn3dmot_2020,Weng-2020-123397,yin_centerbased_2020}.

One line of work in 3D MOT establishes tracks directly from the output of an object detector and forms tracks by connecting detected objects between frames. These approaches can directly use the output of an object detector~\cite{yin_centerbased_2020} or more advanced features including 2D information for every detection~\cite{weng_gnn3dmot_2020, zhang_robust_2019}. In this framework, Weng \etal~\cite{weng_gnn3dmot_2020} are the first to use a graph neural network to estimate the affinity matrix, which is then solved using the Hungarian algorithm. Since this group of trackers does not establish a predictive model for each track, they cannot directly account for missed detections or occlusions and require heuristics for these cases.

Another group of trackers~\cite{chiu_probabilistic_2020,kim_eagermot_2020,Weng-2020-123397} resolves this issue by generating a separate representation of tracks and performs tracking by matching active tracks and detections at each timestep. AB3DMOT~\cite{Weng-2020-123397} uses a Kalman filter~\cite{kalman_new_1960} to represent the track state and matches tracks and detections based on intersection over union (IoU). Chiu \etal~\cite{chiu_probabilistic_2020} extend this approach by matching based on the Mahalanobis distance~\cite{mahalanobis1936generalized} to resolve the issue that object size, orientation and position are on different scales. EagerMOT~\cite{kim_eagermot_2020} uses tracks parameterize in 2D and 3D simultaneously to gain performance from multiple modalities. All of these approaches rely on heuristics to generate new tracks, as track initialization can hardly be learned in a purely offline training approach.

\section{Method}
\label{sec:method}

We model the online 3D MOT problem on a graph, where detections are nodes and the optimal sequences of edges that connect the same objects throughout time need to be found. The resulting core tasks are data association by matching of nodes, track initialization while rejecting false positive detections, interpolation of missed/occluded detections, and termination of old tracks.

Without access to future frames due to the time causal nature in the online setting, all of the aforementioned tasks become challenging. In the case of track initialization, for instance, a new detection in the current frame with no link to a track could be a false positive or the first detection of a new track. And similarly for track termination, where an existing track that is not matched to any detections in the current frame may need to be terminated or may only encounter a missed or occluded object. While these dilemmas could often be resolved when future frames become available over time, online tracking performance is crucial for real-time decision systems since it directly influences the behavior of the system.

To jointly resolve these challenges in a learnable framework, we formulate a graph that merges tracks with their underlying dynamic model and detections into a single representation for online MOT. Based on the detections of the last $T$ frames and the active tracks, a graph is built that represents the possible connections between tracks and detections. Starting with local features at every node and edge, NMP is used to distribute information through the graph and to merge it with the local information at each edge and node during multiple iterations. Finally, edges and nodes are classified as active or inactive.
Based on the active edges that connect track and detection nodes, we formulate an optimization problem for data association. This jointly considers matches between tracks and detections and matches between detections at different timesteps to improve the track stability. Based on the connectivity of the remaining active detection nodes, tracks are initialized.

\begin{figure}[tb]
    \centering
    \includegraphics[width=1.0\linewidth, trim=50 120 120 30, clip]{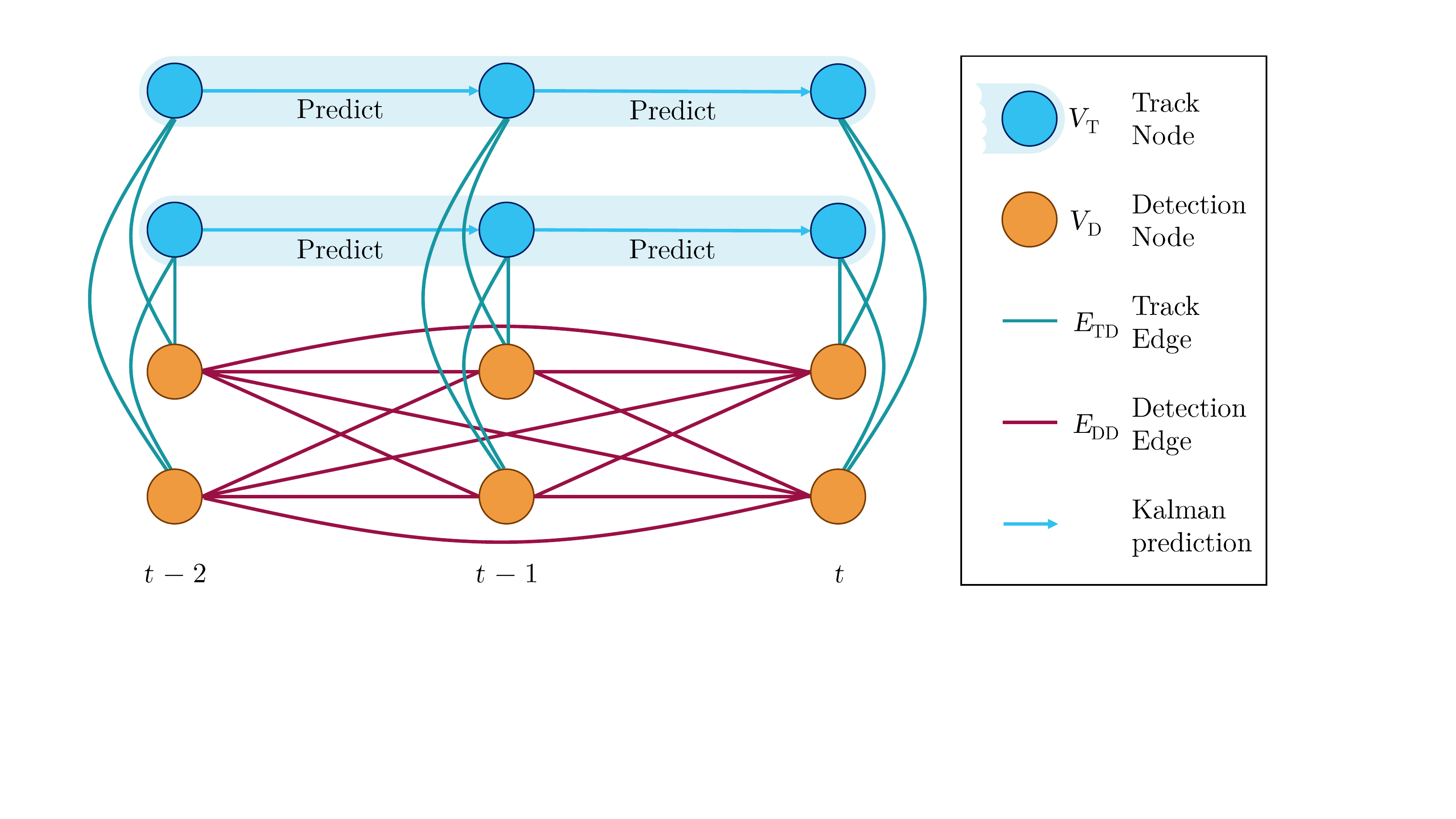}
    \caption{The proposed tracking graph combines tracks, represented by a sequence of track nodes and detections in a single representation. During the NMP iterations, information is exchanged between nodes and edges, and thus, distributed globally throughout the graph.}
    \label{fig:graph}
\end{figure}

\subsection{Graph Representation of Online MOT}
Approaching 3D MOT as tracking by detection can be formulated as finding the set of tracks $\mathcal{T} = \{{T_1}, ..., {T_m}\}$ that underlie the observed set of noisy detections $\mathcal{O} = \{\mathcal{O}_{t_0}, ... \mathcal{O}_{t_T}\}$. We parameterize a track as the state of the underlying Kalman filter and a detection by its estimated parameters such as bounding box, class and velocity. 
To find a robust and time-consistent solution, three tasks need to be solved: 
\begin{enumerate}
    \setlength{\itemsep}{0pt}
    \setlength{\parskip}{0pt}
    \item Assignment of detections to existing track.
    \item Linking of detections across timesteps.
    \item Classification of false positive detections.
\end{enumerate}
While either 1.\ and 2.\ would be sufficient on their own to perform tracking, finding a joint solution promotes stability of the tracks. Furthermore, utilizing a track model is beneficial, since it aggregates information of the complete sequence of matched observations which is required to interpolate missing detections.

The three tracking tasks can be naturally formulated as one joint classification problem on a tracking graph $G = (V_{D}, V_{T}, E_{DD}, E_{TD})$ that is built from detection nodes $V_{D}$ and track nodes $V_{T}$. Detection edges $E_{DD}$ connect pairs of detection nodes at different timesteps and track edges $E_{TD}$ connect track and detection nodes at the same timestep. The complete tracking graph is visualized in Figure \ref{fig:graph}.  
Note that track nodes have sparser connections than detection nodes. They are only connected to the detections at the same timestep and to the neighboring timesteps of the same track. We chose this pattern since connected tracks and detections need to be temporally consistent and the relation between track nodes is determined by the Kalman prediction step. One additional characteristic of track nodes is that nodes that correspond to the same track form a track-subgraph called $G_{T,n}$, which is highlighted with a blue shaded area in Figure \ref{fig:graph}. These subgraphs are important since they share the same state that is linked with a dynamic model. Next, we discuss the types of nodes and edges used in our graph in more detail.

\paragraph{Notation:}
Symbols with subscript $D$ belong to detection nodes and symbols with subscript $T$ to track nodes. Symbols with subscript $DD$ belong to detection edges and symbols with subscript $TD$ to track edges.

Nodes are indexed with integer numbers from the set $\mathcal{I}$ for detection nodes and from $\mathcal{K}$ for track nodes. Edges are referred to by the indices of the connected nodes, \ie, $E_{TD,ki}$ describes a track edge from $V_{T,k}$ to $V_{D,i}$. As the graph is undirected, the notation also holds when the order of the indices is switched. To make our notation easy to read we always use the same index variables. More precisely, the index variables $i,j,m \in \mathcal{I}$ are used to refer to detection node indices and index variables $k,p,q \in \mathcal{K}$ refer to track node indices. The newest timeframe available to the algorithm during online tracking is denoted as $t$ and the timeframe of a specific node is referred to as $t_i$. Finally, tracks are indexed with their track ID $n$.

\paragraph{Detection nodes}
are generated from the detected objects $\mathcal{O}$ and are initialized from the feature $\vct{x}_{D,i}$ containing the position, size, velocity, orientation, one-hot encoded class, detection score, and the distance of the detected object relative to the acquisition vehicle. The position is given in a unified coordinate system which is centered at the mean of all the detections in the graph. The orientation, relative to the same unified coordinate system, is expressed by the angle's $\sin$ and $\cos$.

\paragraph{Track nodes}
represent the state of an active track, i.e., each track generates one track node at every timestep. This groups the track nodes into track-subgraphs. The feature $\vct{x}_{T,k}$ at every track node is defined by the position, size, orientation, and the one-hot encoded class of the tracked object.
The tracks are modeled by a Kalman filter with 11 states corresponding to the position, orientation, size, velocity and angular velocity. Parameters are learned from the training set as proposed by~\cite{chiu_probabilistic_2020}.

\paragraph{Detection edges}
refer to edges between a pair of detection nodes $V_{D,i}, V_{D,j}$ at two different frames $t_i \neq t_j$. They are parameterized by $\vct{x}_{DD,ij}$
containing the frame time difference, position difference, size difference, and the differences in the predicted position assuming constant velocity.
 To reduce the connectivity of the tracking graph, detection edges are only established between detections of the same class and truncated with a threshold on the maximal distance between two nodes. This implicitly corresponds to a constraint on the maximum velocity an object can achieve. Graph truncation makes inference more efficient, track sampling more robust and helps to reduce the strong data imbalance between active and inactive edges.

\paragraph{Track edges}
are connections between a track node $V_{T,k}$ and a detection node $V_{D,i}$ at the same timestep $t_k = t_i$. These edges are modeled with the feature $\vct{x}_{TD,ki}$, where the three entries are the differences in position, size and rotation, respectively.

\paragraph{Classification}
Given the unified tracking graph $G$, the tracking problem is transformed to the following classification tasks:
\begin{enumerate}
    \setlength{\itemsep}{0pt}
    \setlength{\parskip}{0pt}
    \item Classification of active track edges $E_{TD}$.
    \item Classification of active detection edges $E_{DD}$.
    \item Classification of active detection nodes $V_{D}$.
\end{enumerate}
An approach to solving these tasks jointly is presented in the following.

\subsection{Neural Message Passing for Online Tracking}
\label{sec:nmp}

Given only the raw information described in the previous section, classifying edges as active is hard and error-prone.
To generate a good assignment, the network should have access to the global and local information present in the tracking graph.
To archive this exchange of information within the graph, we rely on a graph-NMP network \cite{braso_learning_2020}. Our message passing network for data assigning in the unified tracking graph consists of four stages:

\paragraph{1) Feature embedding:}
The input to the NMP network are embeddings of the raw edge and node features. To generate the 128 dimensional embeddings, the raw features are normalized and subsequently processed with one of four different Multi-Layer Perceptrons (MLP), one for each type of node/edge. This results in the initial features $h_{D,i}^{(0)}, h_{T,k}^{(0)}, h_{DD,ij}^{(0)}, h_{TD,ki}^{(0)}$.

\paragraph{2) Neural message passing:}
Initially, all information contained in the embeddings is local and thus, not sufficient for directly solving the data assignment problem. Therefore, the initial embeddings are updated using multiple iterations of NMP that distribute information throughout the graph. An NMP iteration consists of two steps. First, the edges of the graph are updated based on the features of the connected nodes. In the second step, the features of the nodes are updated based on the features of the connected edges. The networks used to process messages in NMP are shared between all iterations $l = 1,...,L$ of the algorithm. Next, we will describe the NMP iteration for each node and edge type in detail.

\paragraph{Detection Edges} $E_{DD,ij}$ at iteration $l$ are updated with a single MLP $\mathcal{N}_{DD}$ that takes as an input the features of the two connected detection nodes $h_{D,i}^{(l-1)}, h_{D,j}^{(l-1)}$, the current feature of the edge $h_{D,j}^{(l-1)}$ and the initial feature $h_{DD,ij}^{(0)}$
\begin{equation}
h_{DD,ij}^{(l)} = \mathcal{N}_{DD}\left([h_{D,i}^{(l-1)}, h_{D,j}^{(l-1)}, h_{DD,ij}^{(l-1)}, h_{DD,ij}^{(0)}]\right).
\label{eq:det_edge_update}
\end{equation}
Adding the current and initial edge feature to the input vector corresponds to introducing a skip connection into the unrolled algorithm.

\paragraph{Track edges} $E_{TD,ki}$ are updated according to the same principle as detection edges, using information from connected nodes, but with a separately trained MLP $\mathcal{N}_{TD}$. The update rule is given as
\begin{equation}
h_{TD,ki}^{(l)} = \mathcal{N}_{TD}\left([h_{T,k}^{(l-1)}, h_{D,i}^{(l-1)}, h_{TD,ki}^{(l-1)}. h_{TD,ki}^{(0)}]\right).
\label{eq:track_edge_update}
\end{equation}

\paragraph{Detection nodes} are updated with a time-aware node model proposed by Bras\'o \etal~\cite{braso_learning_2020} that we extend with an additional input from connected track edges. Given a fixed detection node $V_{D,i}$, the following messages are generated for every detection edge $E_{DD,ij}$ and tracking edge $E_{TD,ki}$ connected to it,
\begin{align}
m_{D,ij}^{(l)}&=\mathcal{N}^{\text{past}}_D \left([h_{DD,ij}^{(l)}, h_{D,i}^{(l-1)}, h_{D,i}^{(0)}]\right)\; ,\; j\in N^{\text{past}}_i \nonumber\\
m_{D,ij}^{(l)}&=\mathcal{N}^{\text{fut}}_D \left([h_{DD,ij}^{(l)}, h_{D,i}^{(l-1)}, h_{D,i}^{(0)}]\right)\;\; ,\; j\in N^{\text{fut}}_i \label{eq:det_node_messages}\\
m_{D,ki}^{(l)}&=\mathcal{N}^{\text{track}}_D \left([h_{TD,ki}^{(l)}, h_{D,i}^{(l-1)}, h_{D,i}^{(0)}]\right) ,\; k\in N^{\text{track}}_i.\nonumber
\end{align}
The first two message types are time-aware detection messages that consider detection edges to past and future nodes and the third type processes the track edges. The three messages are computed with separate MLPs; $\mathcal{N}^{\text{past}}_D$ is applied to edges $E_{DD,ij}$ that are connected to detection nodes in a frame prior to the considered node. $\mathcal{N}^{\text{fut}}_D$ is the network used to process information on edges $E_{DD,ij}$ that are connected to detection nodes of future frames. Finally, $\mathcal{N}^{\text{track}}_D$ is the network used for track edges. All networks get the current and initial feature of node $V_{D,i}$ as an input to establish skip connections. Note that in the first and last time frame, where no past respectively future edges are available, zero padding is used. 

The messages formed at the incident nodes are aggregated separately for the three types of connections by a symmetric aggregation function $\Phi$
\begin{equation}
\begin{aligned}
&m_{D, i, \text{past}}^{(l)} \hspace{-0.4cm}&&= \Phi \left(\left\{ m_{D,ij}^{(l)}\right\}_{j\in N^{\text{past}}_i}\right)\\
&m_{D, i, \text{fut}}^{(l)} \hspace{-0.4cm}&&= \Phi \left(\left\{ m_{D,ij}^{(l)}\right\}_{j\in N^{\text{fut}}_i}\right)\\
&m_{D, i, \text{track}}^{(l)} \hspace{-0.4cm}&&= \Phi \left(\left\{ m_{D,ki}^{(l)}\right\}_{k\in N^{\text{track}}_i }\right).
\end{aligned}
\label{eq:det_node_agg_messages}
\end{equation}
Aggregation functions commonly used in NMP are summation, the mean or the maximum of all the inputs.
In our implementation, we choose the summation aggregation function.

The node feature is updated with the output of a linear layer, processing the aggregated messages as
\begin{equation}
h_{D,i}^{(l)} = \mathcal{N}_{D}\left(\left[m_{DD, i, \text{past}}^{(l)}, m_{DD, i, \text{fut}}^{(l)}, m_{TD, i, \text{track}}^{(l)}\right]\right).
\label{eq:det_node_update}
\end{equation}

At \textbf{track nodes} only track edges are incident and therefore no separate handling of edges is required. The messages sent from track edges are formed by
\begin{equation}
m_{T,ki}^{(l)}= \mathcal{N}_T\left([h_{TD,ki}^{(l)}, h_{T,k}^{(l-1)}, h_{T,k}^{(0)}]\right),
\label{eq:track_node_messages}
\end{equation}
and accumulated using the aggregation function $\Phi$ as before
\begin{equation}
    m_{T, k}^{(l)}= \Phi \left(\left\{ m_{T,ki}^{(l)}\right\}_{i \in N_k}\right).
\end{equation}
Finally, the message is processed by a single linear layer
\begin{equation}
h_{T,k}^{(l)} = \mathcal{N}_{T}^{'}\left(m_{T, k}^{(l)}\right).
\label{eq:track_node_update}
\end{equation}

These NMP steps are performed for $L$ iterations, which generates a combination of local and global information at every node and edge of the graph.

\paragraph{3) Classification:}
The node and edge features available after performing NMP can be used to classify detection nodes, detection edges, and track edges as active or inactive. Detection nodes need to be classified as active if they are part of a track or initialize a new track and as inactive if they represent a false positive detection. Detection edges and track edges are classified as active if the adjacent nodes represent the same object. For each of the tasks, a separate MLP that takes the final features, $h_{D,i}^{(L)}, h_{DD,ij}^{(L)}$, and $h_{TD,ij}^{(L)}$, is used to estimate the labels
$y_{D,i}$,
$y_{DD,ij}$, and
$y_{TD,ki}$.
The result of the classification stage are three sets. First, the set of active detection node indices
\begin{equation}
    \mathcal{A}_{D} = \left\{i \in \mathcal{I} \;|\; y_{D,i} \geq 0.5\right\}.
\end{equation}
Secondly, the set of active detection edge indices
\begin{equation}
    \mathcal{A}_{DD} = \left\{i,j \in \mathcal{I} \times \mathcal{I} \;|\; y_{DD,ij} \geq 0.5\right\}.
\end{equation}
Finally, the set of active track edge indices
\begin{equation}
    \mathcal{A}_{TD} = \left\{k,i \in \mathcal{K} \times \mathcal{I} \;| \; t_k = t_i \land y_{TD,ki} \geq 0.5\right\}.
\end{equation}

Note that during training, classification is not only performed on the final features $h^{(L)}$ but also during earlier NMP iterations.
This distributes the gradient information more evenly throughout the network and helps to reduce the risk of vanishing gradients.

\paragraph{4) Track update:}
In the last stage of our algorithm, we use the sets of active nodes and edges, to update and terminate existing tracks as well as to initialize new tracks. We achieve this with a greedy approach that maximizes the connectivity of the graph. 

\paragraph{Updates} of tracks are performed by finding the matching detection nodes in the graph for each track and time step. An assignment is a set of detection node indices
\begin{equation}
   \mathcal{F}_n \subset \mathcal{I} : ~|\mathcal{F}_n| \leq T \text{ and } \forall i,j \in \mathcal{F}_n: t_i \neq t_j \text{ if } i \neq j
\end{equation}
from different timesteps. We  define the best assignment as the set of indices corresponding to detection nodes that are 1)~all connected to the track-subgraph $G_{T,n}$ and 2)~have the most active detection edges connecting them.
To find the best assignment for a track $n$, we start with the set of detection node indices that are connected to a track node $V_{T,k}$ through an active track edge.
\begin{equation}
    \mathcal{C}_{D,k}^{node} = \left\{ i \in \mathcal{I} \;| ki \in \mathcal{A}_{TD}\right\}.
\end{equation}
By considering all track nodes of the track-subgraph $G_{T,n}$, the set of detection edge indices connected to a track is defined as
\begin{equation}
    \mathcal{C}_{D,n} = \bigcup_{k\in G_{T,n}} \mathcal{C}_{D,k}^{node}.
\end{equation}
Finally, the set of active detection edge indices between these nodes is derived as
\begin{equation}
    \mathcal{C}_{DD,n} = \left\{ ij \in \mathcal{C}_{D,n} \times \mathcal{C}_{D,n} \;| ij \in \mathcal{A}_{DD}\right\}.
\end{equation}
The quality of the assignment $\Gamma$ representing the optimization problem is the number of detection edges between the assignment nodes that is also present in $\mathcal{C}_{DD,n}$
\begin{equation}
    \Gamma = \left|\left\{\mathcal{F}_n \times \mathcal{F}_n\right\} \cap \mathcal{C}_{DD,n}\right|.
\end{equation}

\begin{figure}[tb]
    \centering
    \includegraphics[width=0.95\linewidth, trim=30 260 560 20, clip]{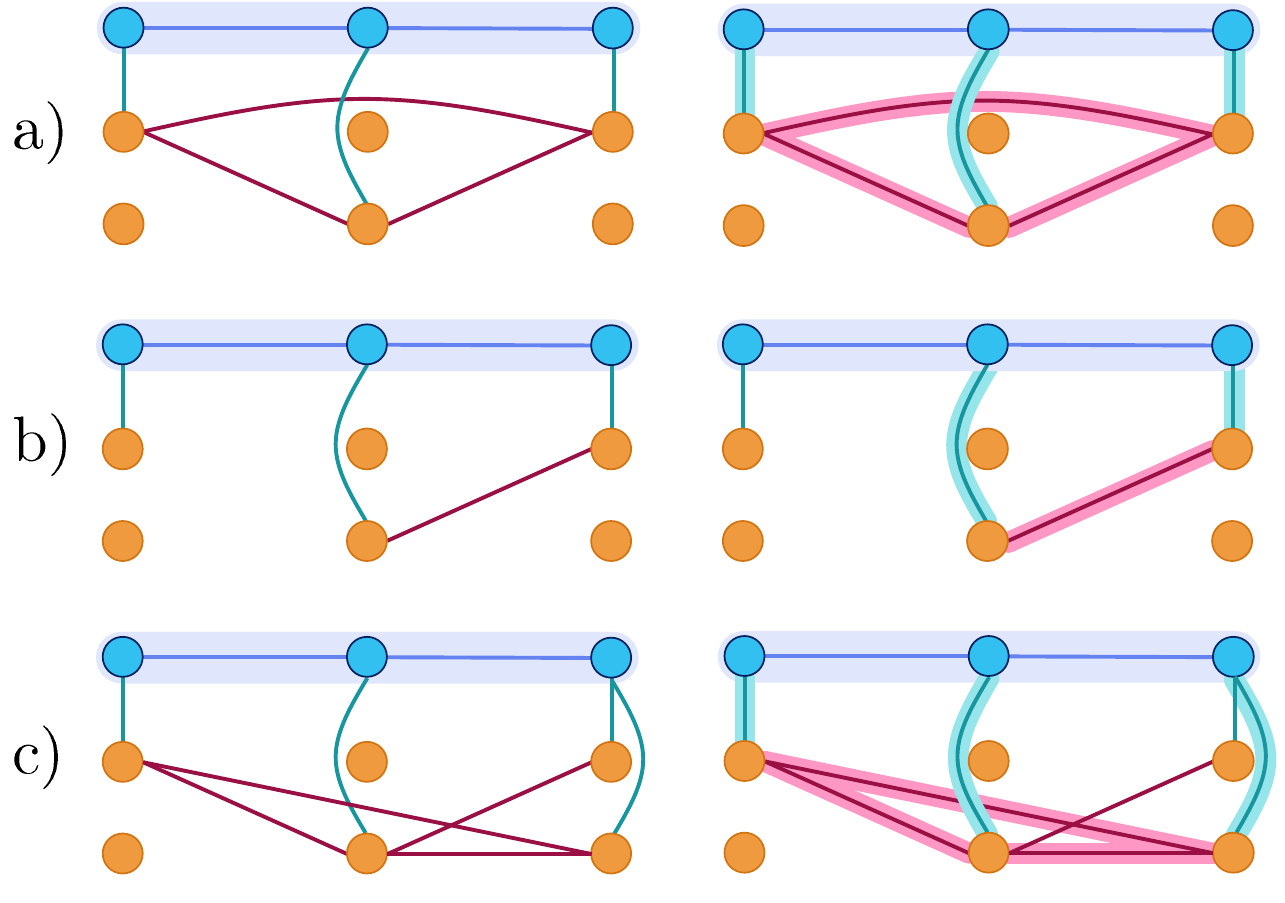}
    \caption{Visualization of different update scenarios, with only active edges in the graph. The graph represents a single track and two detections at each time step. a) Shows the ideal case where a track is matched to one node at every timestep and each detection node is connected with each other. b) Represents the case where a match at one timestep is dropped and the track is only matched to two detection nodes. c) Shows a situation, where the proposed approach is able to decide for the globally best solution, even though two detection nodes have been matched to the track in the last frame.}
    \label{fig:track_update}
\end{figure}

A solution for all tracks is searched with a greedy algorithm,  while never assigning a detection node multiple times. As older tracks are more likely true positive tracks, updating is done by descending age of tracks. If there are multiple solutions with the same cost, we employ the following tie breaking rules. First, solutions with the lowest number of nodes are selected. If this does not make the problem unambiguous, the solution that maximizes the sum of 3D detection scores of the selected detection nodes is chosen. The complete algorithm is provided in the supplementary material and a visualization of this approach is shown in Figure \ref{fig:track_update}.

\paragraph{Termination} of tracks is based on the time since the last update. If a track has not been updated for three timesteps or 1.5s, it is terminated.

\paragraph{Initialization} of tracks takes into account detection nodes and the corresponding detection edges. 
Our approach consists of two steps, split over two consecutive frames.
First, all active detection nodes in the most recent frame that have not been used for a track update are labeled as preliminary tracks. In the next iteration of the complete algorithm, these nodes are in the second to last frame. A full track is generated for each of these nodes that are connected to an unused active detection node in the newest frame by an active detection edge. If multiple active detection edges exist, the edge that connects to the node with the highest detection score is chosen.

\subsection{Training Approach}
\label{sec:training_approach}
When training an online tracker, we face one fundamental challenge, which is the distribution mismatch of track nodes during training and inference. While the track nodes available during training are derived from the ground truth annotations in the dataset, the track nodes encountered during inference are generated by the algorithm itself in a closed loop.

\paragraph{Data augmentation:}
We use data augmentation to make the model more robust against changes in the distribution of tracks and detections as well as to simulate rare scenarios. Although the data naturally contain imperfections such as missed detections and noise on the physical properties of objects, we perform four additional data augmentation steps. Detections are dropped randomly from the graph to simulate missed or occluded detections. Noise is added to the position of the detected objects. 
This allows us to counteract the well-known issue of detector overfitting~\cite{nuscenes2019}, where the detections used for training the tracking algorithm are considerably better than the detections available during inference, as the detector was trained on the same data as the tracker.
To model track termination, all detections assigned to randomly drawn tracks are removed.
Finally, track initialization is simulated by dropping a complete track while keeping the corresponding detection nodes. This ensures that the case of track initialization is encountered often during training.

\paragraph{Two-stage training:}
\label{sec:two_stage}
Data augmentation helps to train a better data association model, however, even with data augmentation, the model does not learn to perform association decisions in a closed loop. To overcome this challenge one could train with fixed length episodes where only the beginning is determined by the ground truth, and for the remaining part, the model data associations and tracks are used to train the model. However, such an approach comes with two issues. First, it is inherently hard to train due to potentially large errors and exploding gradients. Secondly, this approach is computationally costly on large datasets as no precomputed data can be used. Thus, we propose a two-stage training scheme as an alternative that approaches the same challenge. In this setting, a model is trained first on offline data with strong data augmentation. To do so, the results obtained from a LIDAR detection model~\cite{zhu_classbalanced_2019, yin_centerbased_2020} are matched with the annotation data available for the training and validation dataset. The detections matched to tracks are then processed with the Kalman filter model to generate track data for training.

After training the full model on the offline data with data augmentation, the model can be used for inference in an online setting. We run the tracker on the complete training dataset and generate tracks that show a distribution closer to the online-case. This results in a new dataset, which contains the same set of detections as before, but updated tracks.
By retraining the model on this second stage dataset, together with all data augmentation steps used before, considerable performance gains can be accomplished.

\paragraph{Training parameters:}
We train all models with the Adam~\cite{DBLP:journals/corr/KingmaB14} optimizer for four epochs with a batch size of 16 and a learning rate of $0.0005$. Focal loss~\cite{lin_focal_2017} with $\beta = 1$ is used for classification of edges and nodes, weight decay is set to 0.01 and weights are initialized randomly. In all experiments, graphs with $T=3$ timesteps are considered.

\section{Experiments and Results}
All experiments are performed on the publicly available nuScenes dataset~\cite{nuscenes2019} with LIDAR detections only. Scores on the test set are centrally evaluated and results on the validation set are computed with the official developer's kit. NuScenes is known to be more challenging than previous datasets~\cite{weng_gnn3dmot_2020}, thus, providing a suitable platform to test state-of-the-art detection and tracking approaches.
To demonstrate that our method generalizes across significantly different object detectors and provides the same advantages in all scenarios, we perform all experiments with two different object detectors.

\subsection{Detection Data}
To verify the performance of our method with multiple detectors, we choose the two state-of-the-art detectors CenterPoint~\cite{yin_centerbased_2020} and MEGVII~\cite{zhu_classbalanced_2019} that are based on very different techniques. While CenterPoint currently provides the best performance of all publicly available methods, MEGVII is used by many previous methods. We perform all experiments with both detectors and thus, allow for a fair comparison between approaches.

\subsection{MPN Baseline}
\label{sec:baseline}
To show the merit of an explicit graph representation, we implement our method without track nodes and track edges as a baseline. This corresponds to an adaptation of the tracker introduced in~\cite{braso_learning_2020} to the online and 3D MOT setting. In this case, tracks are modeled as a sequence of detections and matching is performed with the classified detection edges and nodes. This method is denoted as MPN-baseline in the following.

\begin{table*}[tbh]
\vspace{3pt}
\begin{center}
\small
\begin{tabular}{l|c|c|cccccc}
Method
&Detections
&Data
&AMOTA\textuparrow
&AMOTP\textdownarrow
&MOTA\textuparrow
&MOTP\textdownarrow
&IDS\textdownarrow
&FRAG\textdownarrow\\
\hline
AB3DMOT\cite{Weng-2020-123397}
&MEGVII\cite{zhu_classbalanced_2019}
&3D
&0.151
&1.501
&0.154
&0.402
&9027
&2557\\
StanfordIPRL\cite{chiu_probabilistic_2020}
&MEGVII\cite{zhu_classbalanced_2019}
&3D
&0.550
&0.798
&0.459
&0.353
&950
&776\\
GNN3DMOT*\cite{Weng-2020-123397}
&-
&2D + 3D
&0.298
&-
&0.235
&-
&-
&-\\
CenterPoint\cite{yin_centerbased_2020}
&CenterPoint\cite{yin_centerbased_2020}
&3D
&0.638
&0.555
&0.537
&\textbf{0.284}
&760
&529\\
CenterPoint-Ensemble*
&CenterPoint Ensemble*
&3D
&0.650
&\textbf{0.535}
&0.536
&0.294
&684
&553\\
Ours
&CenterPoint\cite{yin_centerbased_2020}
&3D
&\textbf{0.656}
&0.620
&\textbf{0.554}
&0.303
&\textbf{288}
&\textbf{371}
\end{tabular}
\end{center}
\caption{Results on the nuScenes test set. Methods marked with asterisk use private detections and thus, no direct comparison is possible. }
\label{table:results_test}
\end{table*}

\subsection{Tracking Results}
The results on the nuScenes test set are shown in Table \ref{table:results_test}. It depicts all competitive LIDAR based methods, which were benchmarked on nuScenes and have at least a preprint available. Our approach achieves an AMOTA score of 0.656, outperforming the state-of-the-art tracker CenterPoint~\cite{yin_centerbased_2020} by 1.8\% using the same set of detections. Compared to CenterPoint-Ensemble, which uses multiple models and an improved set of object detections that are not publicly available, we improve by 0.6\%. Finally, ID switches and track fragmentation are reduced by 58\% and 30\% respectively. This improved track stability can be explained by the integration of the predictive track model into the learning framework.

Our algorithm runs with 12.3\;fps or 81.3\;ms latency on average on an Nvidia TitanXp GPU. As 57.8\;ms of this time is used for graph generation and post-processing and only 23.4\;ms is required for NMP and classification, major gains may be achieved with a more efficient implementation. Further details about the runtime are given in the supplementary material.

Table \ref{table:results_val} shows the results of the current state-of-the-art 3D trackers with two different sets of detections, making them comparable. In this scenario, our approach gains $2.8\%$ AMOTA score compared to CenterPoint~\cite{yin_centerbased_2020} on their own detection data and 3.3\% on the reference MEGVII~\cite{zhu_classbalanced_2019} detections. Again the advantages of using a dedicated model for tracks becomes apparent in the number of ID-switches, which are reduces by 47\% and 43\% using our model on Centerpoint~\cite{yin_centerbased_2020} and MEGVII~\cite{zhu_classbalanced_2019}, respectively.

\begin{table}[tb]
\vspace{3pt}
\begin{center}
\small
\renewcommand\baselinestretch{1.1}\selectfont
\begin{tabular}{l|p{0.9cm}p{0.9cm}p{0.7cm}p{0.5cm}p{0.5cm}}
Method
&AMOTA
&AMOTP
&MOTA
&IDS
&FRAG
\\
\hline
\multicolumn{6}{l}{Detections: MEGVII\cite{zhu_classbalanced_2019}}\\
AB3DMOT\cite{Weng-2020-123397}
&0.509
&0.994
&0.453
&1138
&742\\
StanfordIPRL\cite{chiu_probabilistic_2020}
&0.561
&0.800
&0.483
&679
&606\\
CenterPoint\cite{yin_centerbased_2020}
&0.598
&\textbf{0.682}
&0.504
&462
&462\\
MPN-baseline
&0.514
&0.979
&0.451
&1389
&520\\
Ours
&\textbf{0.631}
&0.762
&\textbf{0.541}
&\textbf{263}
&\textbf{305}\\
\hline
\multicolumn{6}{l}{Detections: CenterPoint\cite{yin_centerbased_2020}}\\
AB3DMOT\cite{Weng-2020-123397}
&0.578
&0.807
&0.514
&1275
&682\\
StanfordIPRL\cite{chiu_probabilistic_2020}
&0.617
&0.984
&0.533
&680
&515\\
CenterPoint\cite{yin_centerbased_2020}
&0.665
&\textbf{0.567}
&0.562
&562
&424\\
MPN-baseline
&0.593
&0.832
&0.514
&1079
&474\\
Ours
&\textbf{0.693}
&0.627
&\textbf{0.602}
&\textbf{262}
&\textbf{332}
\end{tabular}
\end{center}
\caption{Results on the nuScenes validation set. MPNTrack$^{\dag}$ corresponds to the method in~\cite{braso_learning_2020} adapted to the online setting as described in Section~\ref{sec:baseline}.
\label{table:results_val}}
\end{table}

\subsection{Ablation Study}
We evaluate the modules of our tracker in an ablation study shown in Table \ref{table:ablation}. We perform the full study on both sets of detections and for the two training scenarios. The results labeled \textit{online} in Table \ref{table:ablation} refer to our two-stage training pipeline and results labeled \textit{offline} correspond to only training in the first stage of this approach, where no data is generated by the tracker itself. In all cases, inference is performed online.

The results indicate that all implemented modules benefit our method. The highest impact is achieved by propagating information globally using NMP. Next to this, removing information from edges impacts performance for both training approaches. Without node information, the performance drop depends on the dataset. While for Centerpoint the performance drop is severe, it is smaller in the case of MEGVII detections, especially in the offline training case. This may be explained by the quality of detections. While the position information is encoded on nodes and edges, information like object size is only contained on the nodes. Such information has only small variations between different objects and thus, it can only be used effectively if the detection quality is high, as given for CenterPoint.

To remove track nodes, we use the baseline implementation as introduced in Section \ref{sec:baseline}. As only detections are used, this approach does not suffer from a distribution mismatch and two-stage training is neither necessary nor possible. Therefore, while the impact for offline training seems reasonable, the overall impact in the full method is significant. To show the benefit of using detection and track edges jointly for the track update, a naive matching only using track edges in the latest frame is used. This approach performs worse than not using a separate track representation at all and supports our approach of using global information for matching. Finally, focal loss gives a small advantage in all settings and data augmentation helps, especially for offline training. This can be explained, as in the two-stage training, the data distribution is closer to the distribution encountered during inference and thus, less data augmentation is required.

\begin{table}[tb]
\vspace{3pt}
\begin{center}
\small
\begin{tabular}{l|cccc}
&\multicolumn{2}{c}{CenterPoint\cite{yin_centerbased_2020}}
&\multicolumn{2}{c}{MEGVII\cite{zhu_classbalanced_2019}}\\
Method
&online
&offline
&online
&offline
\\
\hline

w/o NMP
&0.427
&0.427
&0.557
&0.499\\

w/o edge features
&0.502
&0.521
&0.460
&0.359
\\

w/o node features
&0.652
&0.587
&0.610
&0.582
\\

w/o track nodes
&(0.593)
&0.593
&(0.582)
&0.582
\\

na\"ive matching
&0.576
&0.427
&0.529
&0.406\\

w/o focal loss
&0.684
&0.647
&0.618
&0.581
\\

w/o data augmentation
&0.688
&0.601
&0.630
&0.538
\\

full pipeline
&0.693
&0.654
&0.631
&0.587\\
\end{tabular}
\end{center}
\caption{Comparative ablation study performed with detections from CenterPoint~\cite{yin_centerbased_2020} and MEGVII~\cite{zhu_classbalanced_2019}. Online refers to the two-stage training introduced in Section \ref{sec:two_stage} and offline to the basic training approach not using self generated data.}
\label{table:ablation}
\end{table}

\section{Conclusion}
We proposed a unified tracking graph representation that combines detections and tracks in one graph, which improves tracking performance and replaces heuristics. We formulated the online tracking tasks as classification problems on the graph and solve them using NMP. To efficiently update tracks, we introduce a method that jointly utilizes matches between all types of nodes. For training, we propose a semi-online training approach that allows us to efficiently train the network for the closed-loop tracking task. Finally, we performed exhaustive numerical studies showing state-of-the-art performance with a drastically reduced number of ID switches. As our proposed method provides a flexible learning based framework, it allows for a wide range of possible extensions and enables the way towards integrating fully learning based track state representations.

\textbf{Acknowledgements}: This work is funded by Toyota Motor Europe via the research project TRACE-Z\"urich.

\FloatBarrier
\clearpage

{\small
\bibliographystyle{ieee_fullname}

}

\FloatBarrier
\clearpage

\appendix

\section{Supplementary Material}
The additional information provided in the supplementary material aims at giving a more thorough insight into some of the technical parts of our paper and to highlight qualitative results. The material contains a description of the chosen network architecture, the algorithm used for updating tracks and additional details on the training approach in Sections \ref{sec:arch}, \ref{sec:update} and \ref{sec:online_training} respectively. Besides this, we provide a runtime analysis in Section \ref{sec:runtime}. Finally, qualitative results are presented in Section \ref{sec:results}, including images and a failure cases analysis.

\subsection{Network Architecture}
\label{sec:arch}
This section provides more details on the separate subnetworks used in the Neural Message Passing (NMP) algorithm. All embedded feature representations are 128 dimensional, which is also the output dimensionality of all networks if not indicated differently. All networks are fully connected with ReLU \cite{Nair:2010:RLU:3104322.3104425} activation functions as non-linearities.

The four encoder networks that process the input features and create the node and edge feature embeddings $h_{D,i}^{(0)}, h_{T,k}^{(0)}, h_{DD,ij}^{(0)}, h_{TD,ki}^{(0)}$, contain two hidden layers with 64 and 128 neurons respectively.

The edge models $\mathcal{N}_{DD}$ and $\mathcal{N}_{TD}$, used during the NMP iterations, both have three hidden layers of size 256, 256 and 128. These operate on the features of the two connected nodes and the previous and first edge feature, enabling skip connections (see Eq. (1) and (2) in the main paper). During the detection node update, the messages from detection edges and track edges are computed with three separate networks $\mathcal{N}^{past}_D, \mathcal{N}^{fut}_D, \mathcal{N}^{track}_D$. Each network has three hidden layers of sizes 256, 256 and 128.
After the aggregation, the 128 dimensional messages from each type of connection $m_{D,ij}^{(l)}, m_{D,ij}^{(l)}, m_{D,ki}^{(l)}$, are concatenated and then jointly processed with a single linear layer and ReLU activation, resulting in the updated feature of dimension 128. 
Track nodes are only updated from the connected track edges and the previous and first node feature. Thus, only a single network $\mathcal{N}_T$ with three hidden layers of size 256, 256 and 128 is used.

Classification of detection edges, track edges and detection nodes is done by three networks of the same dimensions. Each has three hidden layers with 128, 32 and 8 neurons respectively and an output dimension of 1. Using a sigmoid function as the output layer ensures outputs to be within the range $[0, 1]$.

\subsection{Greedy Track Update}
\label{sec:update}
After classifying all relevant nodes and edges in the graph, tracks are updated by assigning detection nodes to them. As described in Section 3.2 of the paper, an assignment $\mathcal{F}_n$ for track $T_n$ is a set of detection node indices at different timesteps. Each of said detection nodes needs to be connected to a track node of the track by an active track edge. The quality $\Gamma$ of the assignment is evaluated as the number of active detection edges between the detection nodes in the assignment. The formalized track update is described with Algorithm \ref{alg:track_update}. In the following, we denote a set of assignments as ${F}$, a single assignment as $\mathcal{F}$ and ab assignment selected for a track $T_n$ as $\widehat{\mathcal{F}}_n$. Furthermore, the total detection score of an assignment
\begin{equation}
    d_\text{total} = \sum_{i \in \mathcal{F}} d_i
    \label{equ:total_det_score}
\end{equation}
is defined as the sum over detection scores of all detection nodes referred to by an assignment.

\begin{algorithm}
 \caption{Track Update. \label{alg:track_update}}\vspace{0.5mm}
    \begin{algorithmic}[1]
    \State Get $\mathcal{A}_{DD}$ and $\mathcal{A}_{TD}$
    \State Sort tracks $T$ by descending age
    \For{$T_n$ in $T$}
        \State Compute all assignments ${F}_n$ for $T_n$
        \State Compute quality $\Gamma$ for each assignment $\mathcal{F}$ from ${F}_n$
        \State Get highest quality $\Gamma_\text{max}$ from ${F}_n$
        \State Store all $\mathcal{F}$ from ${F}_n$ with $\Gamma = \Gamma_{max}$ in $\overline{{F}}_n$
        \If{$|\overline{F}| > 1$}
            \State Compute cardinality $\#$ for each $\mathcal{F}$ from $\overline{{F}}_n$
            \State Get lowest cardinality $\#$ from $\overline{{F}}_n$ as $\#_\text{min}$
            \State Remove all $\mathcal{F}$ from $\overline{{F}}_n$ with $\# > \#_\text{min}$
        \EndIf
        \If{$|\overline{F}| > 1$}
            \State Compute $d_\text{total}$ for each $\mathcal{F}$ from $\overline{{F}}_n$ (Eq. \ref{equ:total_det_score})
            \State Get highest $d_\text{total}$ from $\overline{{F}}_n$ as $d_\text{total, max}$
            \State Remove all $\mathcal{F}$ from $\overline{{F}}_n$ with $d_\text{total} < d_\text{total, max}$
        \EndIf
        \If{$|\overline{F}| > 1$}
            \State Select $\widehat{\mathcal{F}}_n$ from ${F}_n$ randomly
        \Else
            \State Select $\widehat{\mathcal{F}}_n$ as remaining element from ${F}_n$
        \EndIf
        \State Remove edges from $\mathcal{A}_{DD}$ and $\mathcal{A}_{TD}$ that contain any index from $\widehat{\mathcal{F}}_n$
    \EndFor
    \end{algorithmic}
\end{algorithm}

\begin{figure*}[tb!]
    \centering
    \includegraphics[width=\textwidth, trim=0 0 0 0, clip]{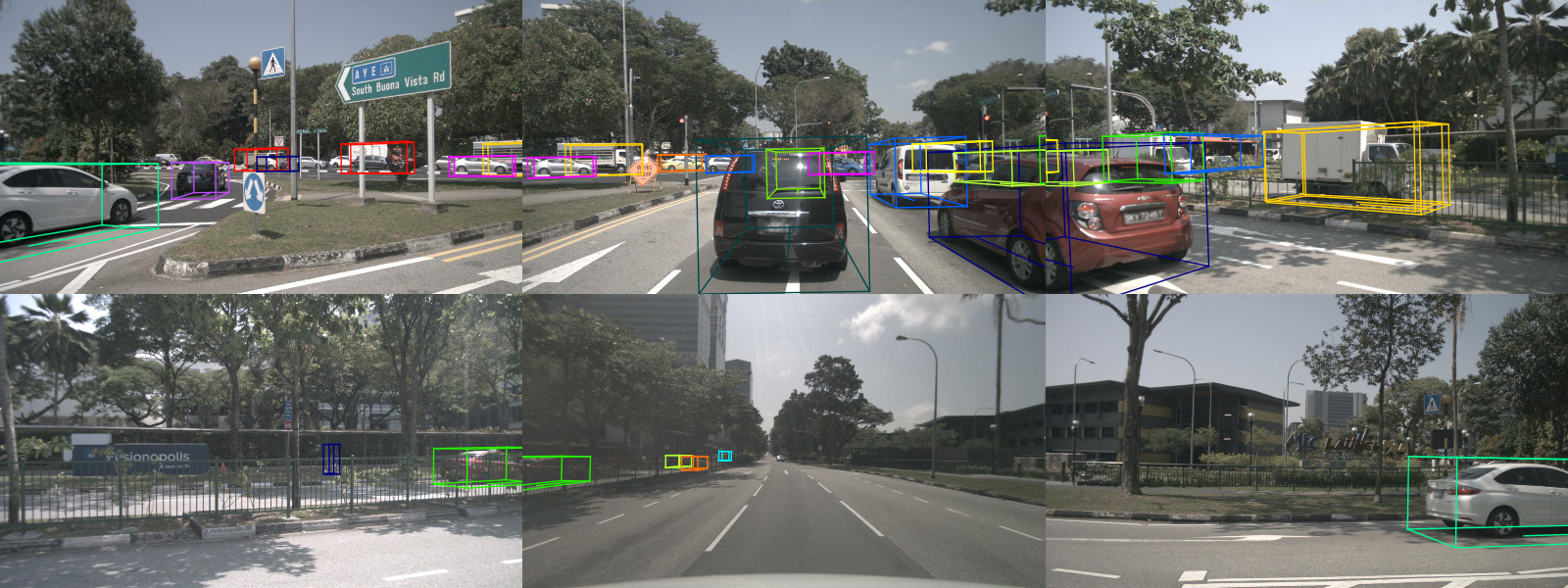}
    \caption{Qualitative results generated with our approach and projected into the $360^\circ$ images.}
    \label{fig:qualitative}
\end{figure*}

\begin{figure}[tb]
    \centering
    \includegraphics[width=\linewidth, trim=0 20 0 0, clip]{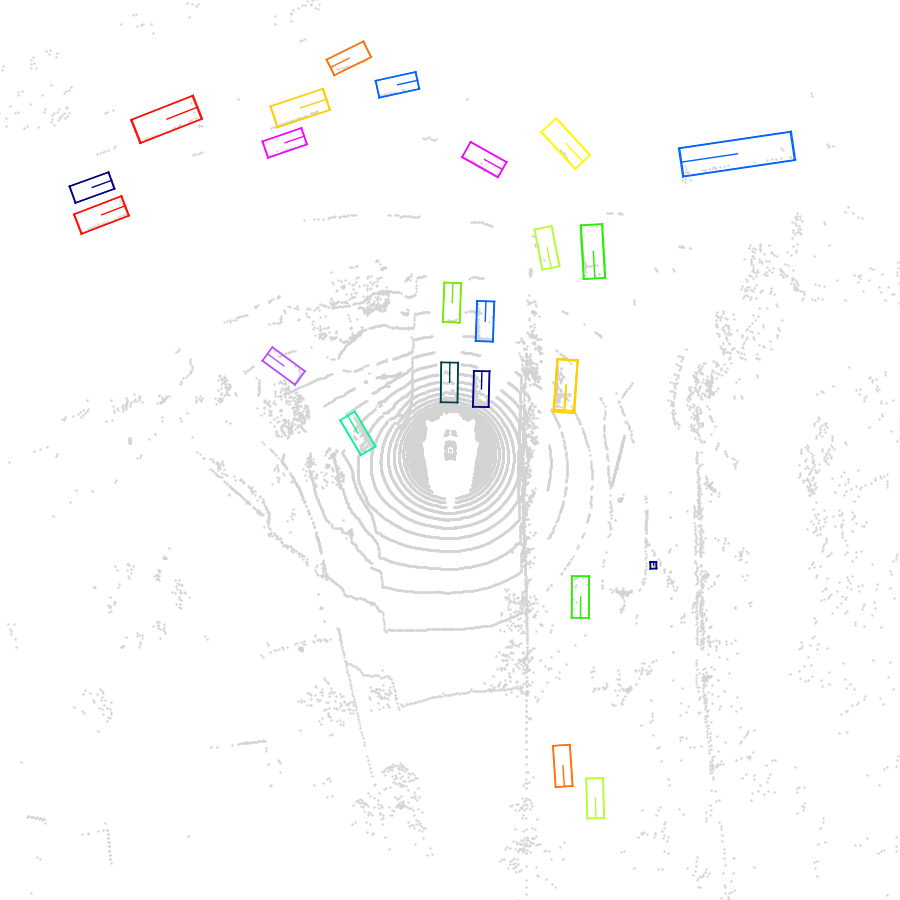}
    \caption{Qualitative results generated with our approach projected into the top-down view containing LIDAR points.}
    \label{fig:qualitative_lidar}
\end{figure}

\subsection{Two-Stage Training}
\label{sec:online_training}
As analyzed in the ablation study, a major performance gain is achieved by using a semi-online two-stage training approach. This refers to training the network with a distribution of tracks that is generated by the network itself. This concept should not be confused with online tracking, i.e. forming tracks without information about future frames, which is a basic assumption for our complete method and all training steps.

For the proposed online-training, in the first stage, a network is trained with data generated from the object detections together with ground truth labels, the step we call offline-training. This way the first set of tracks can be formed, which all have corresponding ground truth tracks. In the second stage, the trained network is used to generate track candidates, where some are close to the ground truth data, but some tracks also come from false positives and should not be matched by the algorithm. This set of tracks cannot easily be represented with the data generated from detections and ground truth data alone, and thus, helps our method to perform better in the closed-loop-setting. 

\subsection{Runtime Analysis}
\label{sec:runtime}
We test the runtime of our algorithm on an Nvidia TitanXp GPU. The runtime of the different components per frame is given in Table~\ref{table:runtime}. \textit{Complete} refers to a complete iteration from generating the tracking graph to returning a list of active tracks. A complete iteration consists of three parts: First the \textit{graph datastructure is generated} from the detections and active tracks. Secondly, \textit{NMP} is performed and the nodes and edges are \textit{classified}. Finally, tracks are continued, terminated or initialized based on the classification results during \textit{post-processing}.

All numbers should be interpreted in the context of the usual sampling rate of 10Hz for LIDAR scanners and the effective sampling rate of 2Hz that is used in the nuScenes \cite{nuscenes2019} dataset. With 12 fps for the complete pipeline, the method can run at the sampling rate of a LIDAR and far above the sampling rate of 2Hz used for tracking in this work.

Optimizing NMP and classification for a production level implementation could yield some reduction in runtime. More importantly, graph generation and post-processing is know to be slow in Python and an implementation in a more efficient programming language may improve the runtime by multiple magnitudes. With around 70\% of the runtime in these components, the real-time capability of the pipeline is further emphasized.

\begin{table}[h]
\vspace{3pt}
\begin{center}
\small
\begin{tabular}{c|cc}
Module
&Latency
&fps\\
\hline
Complete
&81.3\;ms
&12.3\\
\hline
NMP + classification
&23.4\;ms
&42.7\\
Graph generation
&24.9\;ms
&41.7\\
Post-processing
&32.9\;ms
&30.4
\end{tabular}
\end{center}
\caption{Inference time of our algorithm on a Nvidia Titan Xp in the full online setting. All results are measured on the nuScenes\cite{nuscenes2019} validation set.}
\label{table:runtime}
\end{table}

\subsection{Qualitative Results}
\label{sec:results}
Qualitative results achieved with our proposed tracker are shown in Figures \ref{fig:qualitative} and \ref{fig:qualitative_lidar}. The images show the $360^{\circ}$ view  as well as the rendered LIDAR detections of a typical crowded scenario where our tracker outperforms previous methods.

\subsection{Failure Cases}
\label{sec:failure_cases}
During analyzing the qualitative results generated with our tracker, following failure cases have been identified:
(i) Long frame gaps cause scenarios where the time a track is kept active without observations is shorter than the observed occlusion time. While extending this timeframe may further reduce the number of ID-switches, it also increases the number of false positives, harming the overall tracking performance.
(ii) Consistent false positive detections are a general problem for any tracker following the tracking-by-detection paradigm. While our tracker can easily handle isolated false positives due to noise in the detector, cases where e.g. physical structures or reflections are detected cannot be recognized.
(iii) Double objects are generated if the same object is detected multiple times as different types. This behavior can mostly be observed for trucks in our results and should be approached at the detector level \eg with non-maximum suppression\cite{hosang_learning_2017}.

It is important to note that all of these scenarios are also failure cases for existing trackers, including the current state-of-the-art CenterPoint\cite{yin_centerbased_2020}, and are not introduced by our pipeline. Furthermore, while the presented method already reduces the number of such scenarios, further improving the robustness is an interesting line of future work in 3d detection and tracking.

\end{document}